\documentclass[10pt,twocolumn,letterpaper]{article}

\usepackage[pagenumbers]{wacv} % To force page numbers, e.g. for an arXiv version

\usepackage{graphicx}
\usepackage{amsmath}
\usepackage{amssymb}
\usepackage{booktabs}

\usepackage{amsfonts}                                  % blackboard math symbols
\usepackage{nicefrac}                                  % compact symbols for 1/2, etc.
\usepackage{microtype}                                 % microtypography
\usepackage{lipsum}
\usepackage{array} 
\usepackage{tabularx}
\usepackage[table]{xcolor}
\usepackage{multirow}
\usepackage{wrapfig}
\graphicspath{{media/}}     % organize your images and other figures under media/ folder
\usepackage{caption}
\usepackage{subcaption}
\usepackage{adjustbox}

\usepackage{stfloats}

\definecolor{lightest-gray}{gray}{0.97}

\usepackage[pagebackref,breaklinks,colorlinks]{hyperref}

\usepackage[capitalize]{cleveref}
\crefname{section}{Sec.}{Secs.}
\Crefname{section}{Section}{Sections}
\Crefname{table}{Table}{Tables}
\crefname{table}{Tab.}{Tabs.}

\def\wacvPaperID{2236} % *** Enter the WACV Paper ID here
\def\confName{WACV}
\def\confYear{2025}

\begin{document}

\title{VistaFormer: Scalable Vision Transformers for Satellite Image Time Series Segmentation}

\author{
    Ezra MacDonald \\
    University of Victoria \\
    Victoria, BC \\
    \texttt{macdonaldezra@gmail.com}
    \and
    Derek Jacoby \\
    University of Victoria \\
    Victoria, BC \\
    \texttt{derekja@gmail.com}
    \and
    Yvonne Coady \\
    University of Victoria \\
    Victoria, BC \\
    \texttt{ycoady@gmail.com}
}
\maketitle

%%%%%%%%% ABSTRACT
\begin{abstract}
     We introduce VistaFormer, a lightweight Transformer-based model architecture for the semantic segmentation of remote-sensing images. This model uses a multi-scale Transformer-based encoder with a lightweight decoder that aggregates global and local attention captured in the encoder blocks. VistaFormer uses position-free self-attention layers which simplifies the model architecture and removes the need to interpolate temporal and spatial codes, which can reduce model performance when training and testing image resolutions differ. We investigate simple techniques for filtering noisy input signals like clouds and demonstrate that improved model scalability can be achieved by substituting Multi-Head Self-Attention (MHSA) with Neighbourhood Attention (NA). Experiments on the PASTIS and MTLCC crop type segmentation benchmarks show that VistaFormer achieves better performance than comparable models and requires only 8\% of the floating point operations using MHSA and 11\% using NA while also using fewer trainable parameters. VistaFormer with MHSA improves on state-of-the-art mIoU scores by 0.1\% on the PASTIS benchmark and 3\% on the MTLCC benchmark while VistaFormer with NA improves on the MTLCC benchmark by 3.7\%.
\end{abstract}

\section{Introduction}
\label{sec:intro}

Semantic segmentation is a foundational task in computer vision that predicts a class category for each pixel, rather than an image-level prediction \cite{long_fully_2015}. This computer vision task is useful in remote sensing, especially for satellite image time series (SITS) data, where it is necessary to analyze temporal patterns and changes in specific geographical regions. An important application of SITS data is in identifying crop types, since crops undergo phenological events throughout their growth cycle that can be captured in remote sensing imagery \cite{tarasiou_vits_2023, garnot_panoptic_2021}. Accurately identifying crop types has profound for tasks including estimating agricultural yields, monitoring crop health, understanding food security vulnerabilities, creating climate adaptation strategies, and more.

While including additional samples increases the breadth of information in a model's input, it can dramatically increase the dimensions of input data \cite{garnot_panoptic_2021}. Since the earth's surface is covered by more than 60\% clouds \cite{king_spatial_2013, liu_opposing_2023}, many of these additional inputs may be partially or completely obstructed by cloud coverage. The most performant models applied to crop-type segmentation benchmarks are Transformer-based models that apply self-attention along the temporal dimension \cite{cai_revisiting_2023, garnot_panoptic_2021} or both the temporal and spatial dimension \cite{tarasiou_vits_2023}.

This paper introduces VistaFormer, an encoder-decoder model architecture that applies self-attention along the spatial dimension and uses gated convolutions \cite{yu_free-form_2019} to enable downsampling the temporal dimension while rendering profound multi-scale representations. We show that Multi-Head Self-Attention (MHSA) can be substituted with Neighbourhood Attention (NA) \cite{hassani_neighborhood_2023} which dramatically reduces the number of floating point operations required to achieve optimal performance. To verify the performance of VistaFormer, we use two time-series crop-type segmentation benchmarks, namely the MTLCC and PASTIS benchmarks, both of which include few predicted classes and use Sentinel-2 data as inputs. We find that VistaFormer achieves improved overall Accuracy (oA) and mean-Intersection-over-Union (mIoU) scores relative to state-of-the-art performance while using only 8\% of the floating point operations when using MHSA, and 11\% with NA, and also using fewer trainable parameters. These advantages are provided by a model that does not require additional position code interpolation which can reduce performance when resolution in train and test datasets differs \cite{xie_segformer_2021} and make preparing model inputs simpler than in other proposed Transformer-based architectures.

Given the ease of use of the model and its low computational requirements, VistaFormer offers a valuable contribution to advancing time-series deep learning models in a domain aimed at solving some of Earth's most urgent challenges. The code for implementing VistaFormer can be found \href{https://github.com/macdonaldezra/VistaFormer}{here}, and the repository includes links to all data necessary to reproduce the experiments reported in this paper. This paper expands on the research presented in \cite{macdonald_scalable_2024}.

\section{Related Work}

This section reviews research on SITS crop identification, highlighting the progression from traditional machine learning models to advanced neural network approaches, explores the impact of attention mechanisms and Transformer architectures on vision tasks, and discusses the adaptation of these techniques for SITS, comparing our approach to notable models like U-TAE \cite{garnot_panoptic_2021} and TSViT \cite{tarasiou_vits_2023}.

\subsection{SITS Crop Identification}

While some research has been done to use SITS data for identifying general land classes \cite{gomez_optical_2016}, crop type classification has been an especially active area of research \cite{ruswurm_multi-temporal_2018, kussul_deep_2017, ruswurm_temporal_2017, crisostomo_de_castro_filho_rice_2020, garnot_time-space_2019}. Some of the first models to identify crops using SITS data rely on machine learning models such as support vector machines and random forest classifiers, which struggle to learn complex non-linear relationships and often require thoughtfully engineered input features \cite{moskolai_application_2021, zhao_review_2024}. More recent research has demonstrated that models that include neural network layers like RNNs, LSTMs, and convolutions surpass these traditional models in performance \cite{ruswurm_multi-temporal_2018, kussul_deep_2017, ruswurm_temporal_2017, crisostomo_de_castro_filho_rice_2020, garnot_time-space_2019, moskolai_application_2021, zhao_review_2024}. \cite{ruswurm_temporal_2017, crisostomo_de_castro_filho_rice_2020} demonstrate improved crop prediction performance using LSTM-based architectures on Sentinel data while \cite{garnot_time-space_2019} finds that integrating both recurrent and CNN layers improves performance over pure recurrent layer-based architectures. \cite{ruswurm_multi-temporal_2018} uses CGRU \cite{ballas_delving_2016} and CLSTM \cite{shi_convolutional_2015} layers to extract relevant features from raw optical SITS data and shows that these layers are capable of filtering out clouds from inputs. \cite{m_rustowicz_semantic_2019, tarasiou_context-self_2022} show that variations of 3D U-Nets \cite{cicek_3d_2016} have comparable performance for crop segmentation to models that integrate 2D U-Nets \cite{ronneberger_u-net_2015} and recurrent layers.

\subsection{Attention \& Transformers in Vision}

While recurrent layers excel at learning deep representations of sequences, they struggle to process data in parallel and are challenged with learning long-range dependencies. The introduction of attention \cite{bahdanau_neural_2014} and the subsequent introduction of the Transformer \cite{vaswani_attention_2017} architecture, improved on this layer by introducing self-attention mechanisms that enable parallel processing of global sequences and capturing long-range dependencies more effectively, enhancing both computational efficiency and the ability to understand complex patterns in data. While self-attention is highly parallelizable, its computational complexity scales quadratically with the size of the input, making encoding images as a raw sequence of pixels prohibitive for most images. ViT \cite{dosovitskiy_image_2020} introduced the first pure Transformer-based model that achieved state-of-the-art performance in image classification. This model reduced the computational complexity of applying the Transformer to vision by encoding images into patches and treating each patch as a sequence of tokens.

For dense prediction tasks, PVT \cite{wang_pyramid_2021} introduced a pyramid structure-based pure self-attention backbone that outperformed comparable CNN-based architectures. PVT was then improved on by models like Swin \cite{liu_swin_2021, liu_swin_2022}, Twins \cite{chu_twins_2021}, and CoaT \cite{xu_co-scale_2021} that removed fixed size position embeddings to enhance local feature representations and improve model results on dense prediction. \cite{xie_segformer_2021} introduced SegFormer, a more efficient alternative, that among other things introduced a purely data-driven position encoding layer using $3 \times 3$ depth-wise convolutions in the MLP layer of the Transformer. While more recent model architectures like Mask2Former \cite{cheng_masked-attention_2022} and I-JEPA \cite{assran_self-supervised_2023} share structural similarities with the original Transformer architecture, such as employing self-attention mechanisms to process and compare different parts of the input data, the simplicity and effectiveness of the self-attention layer in the Transformer makes it ideal for constructing models that minimize floating point operations and parameter complexity.

\subsection{SITS for Transformers}

Previous work introduced U-TAE \cite{garnot_panoptic_2021} which uses a U-Net architecture with a temporal attention mask that is only computed for the lowest resolution layer and is then upsampled to higher resolution embeddings. These masks are used to collapse the temporal dimension along with a 1D convolution to produce a single map per resolution. We differ from U-TAE \cite{garnot_panoptic_2021} most notably by downsampling both spatial and temporal dimensions after the first encoder layer to reduce floating point operations, and by computing spatial attention in each encoder block.

TSViT \cite{tarasiou_vits_2023} proposes an architecture inspired by ViT \cite{dosovitskiy_image_2020}, that uses input dates to encode temporal positions and uses separate self-attention Transformer layers for computing attention weights along temporal and spatial dimensions. This architecture is effective but computationally expensive in terms of floating point operations since it does not downsample inputs and computes attention on time and space sequences separately. TSViT \cite{tarasiou_vits_2023} optimized model performance by encoding temporal positions using dates of the model by encoding also encodes temporal positions using dates, which does not accommodate integrating additional data sources like radar and requires additional data pre-processing for inputs.

Most recently, \cite{cai_revisiting_2023} introduced a model architecture that computes the similarity between a temporal context cluster and temporal input features. The temporal module is used to wrap a 2D segmentation model, allowing for enhanced model flexibility. The pre-trained model in their experiments holds state-of-the-art performance in terms of mIoU for crop-class segmentation on the PASTIS and MTLCC benchmarks used for our experiments. We compare our model's performance instead to models that have a similar number of trainable parameters, achieve their performance from randomized weights, and report on both mIoU and oA scores as these benchmarks have significant class imbalances.
\vspace{-2mm}

\section{Methods}

In SITS semantic segmentation tasks, we are given an input $\mathbf{X} \in \mathbb{R}^{C \times T \times H \times W}$ and output $\mathbf{Y} \in \mathbb{R}^{K \times H \times W}$ where $C$ denotes input channels, $H$ and $W$ indicate input dimensions, $T$ the number of samples, and $K$ defines predicted classes.

\begin{figure*}[ht]
    \centering
    \includegraphics[scale=0.8]{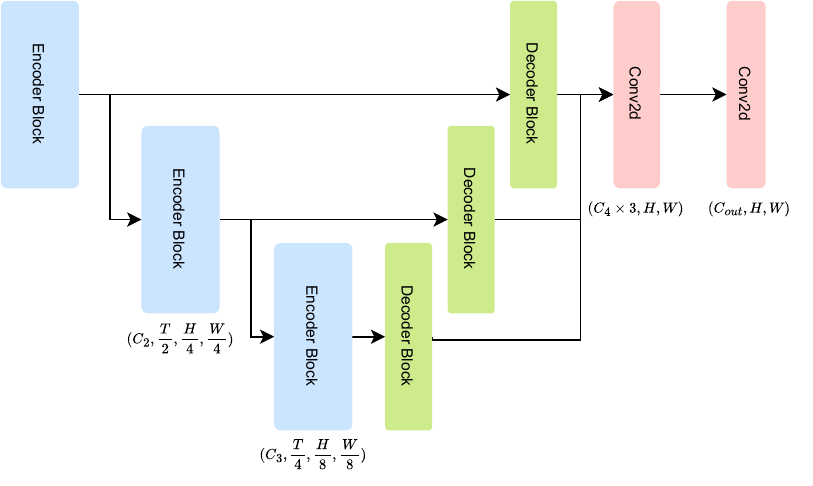}
    \vspace*{-3mm}
    \caption{The VistaFormer model architecture uses a three-layer encoder-decoder architecture where the encoder blocks downsample inputs and computes self-attention while the decoder blocks are comprised of lightweight upsampling layers that unify features from the encoder outputs to generate dense predictions.}
    \label{fig:vistaformer}
\end{figure*}

\begin{figure*}[ht]
    \centering
    \begin{subfigure}[b]{.25\textwidth}
        \centering
        \includegraphics[width=0.95\linewidth]{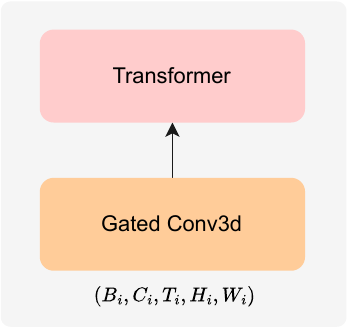}
        \caption{}
        \label{fig:encoder}
    \end{subfigure}
    \begin{subfigure}[b]{.25\textwidth}
        \centering
        \includegraphics[width=0.95\linewidth]{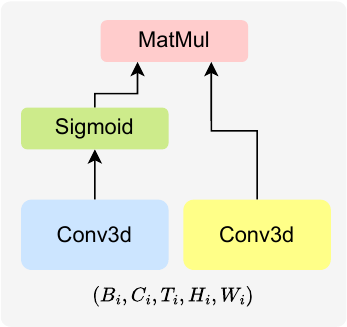}
        \caption{}
        \label{fig:gated_conv}
    \end{subfigure}
    \begin{subfigure}[b]{.25\textwidth}
        \centering
        \includegraphics[width=0.95\linewidth]{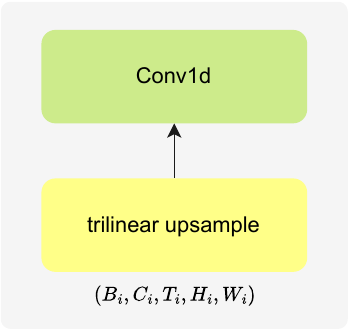}
        \caption{}
        \label{fig:decoder}
    \end{subfigure}
    \caption{\textbf{(a)} Each encoder block downsamples inputs using gated convolutions to reduce atmospheric distortions, reshapes them into sequences of tokens, and processes them through self-attention Transformer layers. \textbf{(b)} The use of gated convolutions implemented here enhances the model's resilience to obstructions like clouds present in input samples. \textbf{(c)} The decoder block uses trilinear upsampling and a 1D convolution to extract features and align embedding dimensions producing a dense prediction.}
    \vspace*{-3mm}
\end{figure*}

Here we introduce VistaFormer, a Transformer-based model designed to take a careful view from a distance, using a simple model architecture to output a dense prediction. We propose a three-layer encoder-decoder architecture, as shown in Figure \ref{fig:vistaformer}, where each encoder block downsamples inputs using gated convolutions and computes self-attention on each of the downsampled inputs. Each of the Transformer layers uses a lightweight depth-wise convolution to encode position information similar to the Transformer blocks used in SegFormer \cite{xie_segformer_2021}. The decoder blocks apply trilinear interpolation to increase the dimensions of each multi-scale representation and use a 1D convolution to collapse the temporal dimension and unify the embedding dimension of each encoder block.
\vspace{-2mm}

\subsection{Encoder}

Each encoder layer, as can be seen in Figure \ref{fig:encoder}, is structured such that inputs are downsampled using 3D convolution layers, reshaped to treat each pixel as a token in a sequence, and then fed through Transformer blocks.
\vspace{-2mm}

\subsubsection{Downsampling} We apply non-overlapping 3D convolutions to each input at each encoder layer and find that using a simplified variation of gated convolutions \cite{liu_image_2018} provides modest performance improvements. VistaFormer uses a gated convolution on input $x$ where $\phi$ denotes convolution and $\sigma$ denotes the sigmoid function:

\vspace{-2mm}
\begin{equation}
    m = \phi_l(x) \odot \sigma(\phi_m(x)) \\
\end{equation}
\vspace{-2mm}

This deviates from the gated convolution presented in \cite{liu_image_2018} in that we do not apply an activation function on $\phi_l(x)$, opting to use $\sigma(\phi_m(x))$ to scale convolution outputs and allow for increased variability in input data. This convolution mechanism was introduced for image in-painting to ignore irrelevant pixels in an input image while computing convolution outputs \cite{liu_image_2018}. We find this downsampling architecture is similarly suitable for cases where input images may contain visual obstructions based on the merits of this model's performance while using gated convolutions that have not been pre-trained to mask out clouds and other atmospheric distortions as seen in Table \ref{tab:ablation}. Given that the applied context for this model is for datasets where individual pixels account for a considerable area, we carefully downsample inputs along spatial dimensions and do not downsample $T$ for the first layer of the encoder block.

\subsubsection{Self-Attention} 

The main computational bottleneck of the encoder is the self-attention Transformer layer. In Multi-Head Self-Attention (MHSA), each of the heads $Q$, $K$, and $V$ have the dimensions $N \times C$, where $N = H \times W$ is the length of the sequence and the self-attention is computed as:

\vspace{-4mm}
\begin{equation}
    \mbox{Attention}(Q, K, V) = \mbox{Softmax}(\dfrac{QK^T}{\sqrt{d_{head}}})V \\
\end{equation}
\vspace{-4mm}

The computational complexity of this process is $O(N^2)$ which presents computational challenges when the input dimensions increase or as the number of $T$ samples increases. By substituting MHSA with 2D NA, which computes attention in a localized neighbourhood around each token of size $k$, we can improve the scalability of VistaFormer by using a neighbourhood $k$ smaller than $N$ \cite{hassani_neighborhood_2023}. We provide an implementation of VistaFormer which uses MHSA and a separate implementation using NA that scales better as the input dimensions increase.

\begin{table}[ht!]
    \label{tab:computational_cost}
    \centering
    \footnotesize{
    \begin{tabular}{ccc}
        \toprule
        \textbf{Module} & \textbf{FLOPs} &  \textbf{Memory} \\
        \midrule \addlinespace[0.3em]
        Self-Attn & $3HWC^2 + 2H^2W^2C$ & $3C^2 + H^2W^2$ \\
        Neighbourhood Attn & $3HWC^2 + 2HWCK^2$ & $3C^2 + HWK^2$ \\
        3D Convolution & $HWC^2K^3$ & $C^2K^3$ \\ 
        \bottomrule
    \end{tabular}}
    \caption[Computation Cost Analysis]{The complexity analysis above provided from \cite{hassani_neighborhood_2023} uses single-head self-attention for simplicity. Note that in our case we increment the FLOPs used for our purposes by a multiple of $T$ since we compute self-attention for every sample $T$.}
    \vspace{-2mm}
\end{table}

\subsubsection{FeedForward Network}

Models like ViT and TSViT use positional encoding (PE) to introduce location information, which fixes the resolution of positional encodings. This can result in reduced performance when the test resolution differs from the training resolution \cite{xie_segformer_2021}. To address this issue and simplify the model implementation, we use a 3 $\times$ 3 depth-wise convolution directly in the feed-forward network (FFN) which was shown to be sufficient to provide positional information for Transformers in \cite{xie_segformer_2021}. To compute the output from the FFN layer we have:

\vspace{-4mm}
\small{
\begin{equation}
    \mbox{FFN}(x) = \mbox{Linear}(\mbox{GELU}(\mbox{Conv3d}_{3 \times 3}(\mbox{Linear}(x)))) + x
\end{equation}
}
\vspace{-4mm}

Observe that GELU \cite{hendrycks_gaussian_2023} is a commonly used activation function in FFN and the Conv3d$_{3 \times 3}$ layer is a depth-wise convolution used to capture positional information. The Conv3d layer is used as it captures position information efficiently since it applies spatial filtering independently to each channel, preserving channel-specific spatial details while reducing computational complexity compared to standard convolutions. Since we downsample the temporal dimension after the first encoder layer, we use a 3D convolution to encode position information instead of a 2D convolution as used in \cite{xie_segformer_2021}.

\subsection{Decoder}

VistaFormer uses a lightweight decoder consisting of an upsampling layer, a 1D convolution to collapse the temporal dimension and ensure the embedding dimensions for encoder layers are consistent, and a 2D convolution to output a mask prediction. We first apply trilinear interpolation to the outputs of each encoder block as this allows for including temporal and spatial information in the upsampled blocks. A 1D convolution is applied to each of the upsampled blocks to output a fixed embedding dimension of size $C$ and collapse the temporal dimension $T$. To combine the multi-scale feature representations, we concatenate each of these layers and use a 2D convolution to output a predicted mask.

\vspace{-4mm}
\begin{equation}
\begin{split}
    U_i &= \mbox{Upsample}(W, H)(E_i), \forall i \\
    F_i &= \mbox{Conv1d}(C_{U_i},C)(U_i), \forall i \\
    G &= \mbox{Concat}(F_i), \forall i \\
    Y &= \mbox{Conv2d}(C, N_{cls})(G)
\end{split}
\end{equation}
\vspace{-2mm}

\noindent where Conv1d$(C_{in}, C_{out})$ and Conv2d$(C_{in}, C_{out})$ denotes the respective convolution layers, $C_{in}$ denotes the input embedding dimension, and $C_{out}$ represents the output embedding dimension. Observe also that the Upsample layer uses trilinear interpolation where $W$ and $H$ correspond to the input dimensions given by $H$ and $W$.

\begin{table*}[ht!]
    \centering
    \resizebox{\textwidth}{!}{
    \tiny
    \begin{tabular}{ccccccc}
        \toprule
        Encoder Layer & Embed Dim &  Patch & Stride & Transformer Layers & Attention Heads & MLP Mult \\
        \hline \addlinespace[0.3em]
        $E_1$ & 32 & (1, 2, 2) & (1, 2, 2) & 2 & 2 & 4 \\
        $E_2$ & 64 & (2, 2, 2) & (2, 2, 2) & 2 & 4 & 4 \\
        $E_3$ & 128 & (2, 2, 2) & (2, 2, 2) & 2 & 8 & 4 \\
        \bottomrule
    \end{tabular}
    }
    \caption{Configuration values for the VistaFormer encoder module where $E_i$ corresponds to a given encoder block.}
    \label{tab:encoder_config}
\end{table*}
\vspace{-2mm}

\section{Experiments}

We now evaluate the performance of the VistaFormer models on two crop-type segmentation benchmarks, PASTIS and MTLCC. In Section \ref{sec:results} we report on the comparison of VistaFormer models to the current state-of-the-art performance, while in Section \ref{sec:ablations} we provide an overview of the performance of some ablations performed on the VistaFormer model that uses MHSA. Finally, we show VistaFormer's scalability relative to U-TAE and TSViT models as input and temporal dimensions vary in Section \ref{sec:scalability}.

\subsection{Datasets}

To evaluate VistaFormer, we use the MTLCC \cite{ruswurm_multi-temporal_2018} and optical PASTIS \cite{garnot_panoptic_2021} semantic segmentation benchmarks. The datasets we chose for evaluating the performance of our model have the following similar noteworthy characteristics (a) they include samples that are obstructed by cloud coverage (in some cases multiple images are entirely covered by clouds), (b) they are both imbalanced datasets with many of the classes accounting for a tiny percentage of the overall pixels, and (c) they include a large number of background pixels that may or may not be easily confused with crop class pixels.

\subsubsection{MTLCC} The MTLCC \cite{ruswurm_multi-temporal_2018} dataset covers an area of 102km $\times$ 42km north of Munich, Germany and includes 17 crop classes along with an unknown class that accounts for 39.91\% of pixels. The dataset includes 13 Sentinel-2 bands split into $24 \times 24$ pixels for the highest resolution bands and we up-sample the lower resolution bands using bilinear interpolation to match the dimensions of the highest resolution bands. The dataset includes samples for 2016 which includes 46 samples and 2017 which includes 52 samples. We use the splits provided in the original study for a direct model comparison which has 27k training samples, 8.5k validation samples, and 8.4k test samples. In keeping with the evaluation criteria used in \cite{tarasiou_vits_2023}, we use 2016 for train, validation, and test datasets. However, we deviate from results reporting from \cite{tarasiou_vits_2023, ruswurm_multi-temporal_2018}, in that we record model results both when the unknown class is included and not included. Not including unknown/background classes during training ensures the model is not penalized for making false predictions in that given area, ensuring the resulting model is more likely to predict false positives, making the model unreliable for predicting realistic boundaries for a class in most applied contexts \cite{yang_sparse_2022, zheng_foreground-aware_2020}. Note that the background class in this benchmark accounts for 43.2\% of the overall pixels while 13 of the remaining 17 crop classes account for just 13.57\% of the overall pixels. 

\subsubsection{PASTIS} The PASTIS \cite{garnot_panoptic_2021} dataset spans over 4,000 km${^2}$ with images taken from four regions in France. Each sequence of images includes 10 Sentinel-2 bands split into $128 \times 128$ pixels and includes between 38 and 61 observations taken between September 2018 and November 2019 \cite{garnot_panoptic_2021}. The dataset includes 2,433 samples that are split into 5 folds where for each split, three folds are used for training; one fold is used for validation; and the remaining one fold is used for testing. There are 5 combinations of splits used for measuring model performance on the dataset to ensure that each of the splits can be independently used as the test dataset to better ensure the model generalizes well. The dataset includes 20 classes, with 18 crop types, a background (or non-crop) class, and a void class which includes either only partial crop class pixels or crop types the authors were unable to confidently identify. The void label is ignored during loss though the background label is included during training and inference results, as specified in \cite{garnot_panoptic_2021}. Note that the background class in this benchmark dataset accounts for 39.91\% of the overall pixels while 15 of the remaining 18 classes account for 13.86\% of the overall pixels.

\subsection{Implementation Details \label{sec:impl_details}}

For both datasets, we train VistaFormer using the weighted Adam optimizer \cite{loshchilov_decoupled_2017} using $\beta_1 = 0.9$ and $\beta_2 = 0.999$ as the coefficients for computing running averages of the gradient and it's square. We use a one-cycle learning rate scheduler that starts with a learning rate of 0.0004 and increases to a max learning rate of 0.01 after the first 10\% of training and is then reduced to a final learning rate of 0.001 in the last epoch. For both datasets, we use a dropout and drop path of 17.5\% respectively and use cross-entropy loss for the loss function as in \cite{tarasiou_vits_2023, garnot_panoptic_2021}. We found using this higher learning rate and learning rate schedule to outperform lower learning rates for both the max learning rate and the scheduled values. For each input, we normalize using techniques detailed in the original papers \cite{garnot_panoptic_2021, ruswurm_multi-temporal_2018} and apply flip and 90$^{\circ}$ rotate transformations for inputs during training with 50\% likelihood of applying the respective transformation. The models were trained using distributed training on compute nodes with 8 CPUs, 100GB of memory, and two Tesla V100 GPUs for roughly 8-12 hours.

We trained on the PASTIS dataset with a batch size of 32 and a maximum sequence length of 60, and height and width of 32, while for the MTLCC dataset, we used a batch size of 16 and a maximum sequence length of 46 and a provided input height and width of 24. Given that the model uses 3D convolutional layers for upsampling and downsampling which contribute significantly to the number of trainable parameters (relative to our model size); decreasing the input sequence length results in a considerably smaller model. For the MTLCC dataset, we found that decreasing the sequence length from 60 to 46 reduced the number of trainable parameters by 13\%. Reducing the sequence dimension for the MTLCC dataset was done in keeping with the sequence length used in \cite{tarasiou_vits_2023} and to reduce the number of blank images included with each sample.

Given the small dimensions of the input images for the datasets used during experimentation and the downsampling rate selected for each encoder level, we found that a model architecture with three input layers outperformed other model architectures that included fewer pairs of encoder-decoder blocks. We also found that the selected batch sizes for both the MTLCC and PASTIS benchmarks were optimal relative to smaller or larger batch sizes. The configurations used for the Encoder are given in Table \ref{tab:encoder_config}, while for the decoder, the unique configuration we used for our model was to use 64 output channels for each of the 1D convolution layers which downsample $T$. We found the attention head dimension outperformed smaller or larger sizes and the selected embedding dimension outperformed larger embedding dimensions at each layer.

For the implementation of VistaFormer that uses 2D NA we chose to use a neighbourhood size $k$ of 13 to increase the spatial extent used for computing self-attention, which is slightly above the window of size 7 used in the experiments for \cite{hassani_neighborhood_2023}. We also deviate from the configurations detailed in Table \ref{tab:encoder_config} by using 1, 2, and 4 attention heads respectively in the encoder blocks.

\begin{table*}[ht]
    \centering
    \vspace{1mm}
    \resizebox{\textwidth}{!}{
    \begin{tabular}{@{\extracolsep{\fill}}ccccc}
            \toprule
            \multicolumn{3}{c}{\ } & \multicolumn{1}{c}{PASTIS} & \multicolumn{1}{c}{MTLCC (2016)} \\ [1ex]
            \cmidrule(r){4-5}
            Model &  GFLOPs & Model Params (m) & oA / mIoU & oA / mIoU \\
            \midrule
            FPN + ConvLSTM \cite{chamorro_martinez_fully_2021} & 282.56 & 1.15 & 81.6 / 57.1 & 91.8 / 73.7 \\
            UNet + ConvLSTM \cite{m_rustowicz_semantic_2019} & 24.52 & 1.52 & 82.1 / 57.8 & 92.9 / 76.2 \\
            UNet-3D \cite{cicek_3d_2016} & 48.63 & 1.55 &  81.3 / 58.4 & 92.4 / 75.2 \\
            U-TAE \cite{garnot_panoptic_2021} & 23.06 & 1.09 & 83.2 / 63.1 & 93.1 / 77.1 \\
            TSViT \cite{tarasiou_vits_2023} & 91.88 & 1.67 & 83.4 / 65.4* & 95.0 / 84.8 \\
 \textbf{VistaFormer Neighbourhood (ours)} & \textbf{9.82} & \textbf{1.13} & \textbf{83.7 $\pm$ 0.2 / 65.3 $\pm$ 0.3} & \begin{tabular}{@{}c@{}} \textbf{96.1 $\pm$ 0.03 / 88.5 $\pm$ 0.05} \\ \textbf{(90.5 $\pm$ 0.08 / 79 $\pm$ 0.16)} \end{tabular} \\
            \textbf{VistaFormer (ours)} & \textbf{7.7} & \textbf{1.25} & \textbf{84.0 $\pm$ 0.1 / 65.5 $\pm$ 0.1} & \begin{tabular}{@{}c@{}} \textbf{95.9 $\pm$ 0.14 / 87.8 $\pm$ 0.5} \\ \textbf{(90.4 $\pm$ 0.1 / 78.7 $\pm$ 0.3)} \end{tabular} \\
            \bottomrule
    \end{tabular}
    }
    \caption[VistaFormer Comparison with SOTA]{Comparison with comparable models on semantic segmentation. Results for PASTIS are reported by computing the average performance across all five folds of the dataset provided in all five folds for comparison with \cite{garnot_panoptic_2021}. For MTLCC, we report results that exclude the unknown class in training and testing in keeping with \cite{tarasiou_vits_2023, ruswurm_multi-temporal_2018} and results including the background/unknown class in parenthesis. Note that results marked with an asterisk for PASTIS were trained using the PASTIS dataset with a height and width of 24.}
    \label{tab:model_results}
\end{table*}

\begin{figure*}[b]
    \centering
    \includegraphics[scale=0.9]{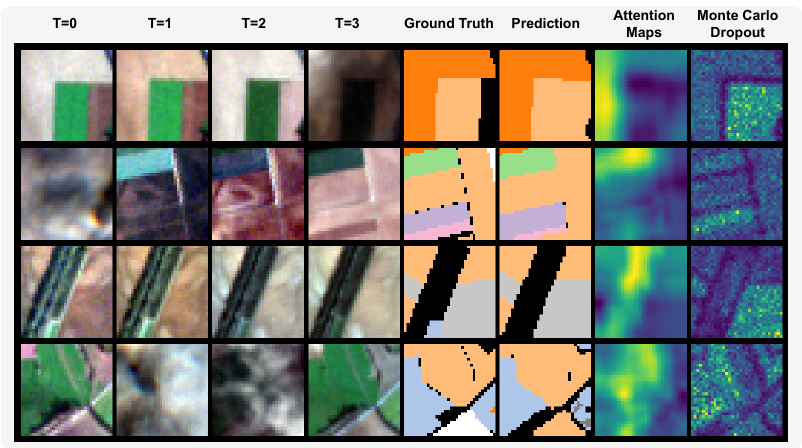}
    \caption[VistaFormer Sample Predictions]{VistaFormer sample semantic segmentation predictions on the PASTIS benchmark. Under titles $T=0, ..., 3$, we show samples of input RGB channels and include these alongside ground truth annotations, model predictions, attention maps, and Monte Carlo dropout \cite{gal_dropout_2016} predictions to measure the uncertainty of model predictions. We use the dropout settings used during training for Monte Carlo Dropout and the outputs reflect the model certainty measure over 10 iterations.}
    \label{fig:predictions}
\end{figure*}

\subsection{Results \label{sec:results}} 

Performance of the model is measured using the mean Intersection over Union (mIoU) score, which computes the averages of the IoU score for each class and the overall Accuracy (oA), which calculates the accuracy summed over all predicted pixels. We find that both VistaFormer models outperform TSViT, the current state-of-the-art model, while using roughly 8\% of the floating point operations for VistaFormer with MHSA and 11\% for VistaFormer using NA. On average, VistaFormer with NA was outperformed by TSViT in the PASTIS benchmark by 0.1\%, though our model improved on the oA score by 0.3\% and improved on the MTLCC score of TSViT by 1.1\% in terms of oA and 3.7\% for mIoU. These metrics were chosen per the metrics used in TSViT \cite{tarasiou_vits_2023} and U-TAE \cite{garnot_panoptic_2021}. Including oA as well as mIoU in a semantic segmentation task with few classes and many pixels labelled as background or unknown provides a straightforward measure of the model's performance across all pixels, ensuring the model effectively distinguishes between relevant and irrelevant regions and maintains performance across all predicted classes. For our model, we report the mean and standard deviation performance over three trials for both benchmarks. GFLOPs are estimated using the FVCore library using an input shape with B=4, T=60, C=10, H=32, and W=32.

\begin{table*}[t]
    \centering
    \resizebox{\textwidth}{!}{
        \begin{tabular}{@{\extracolsep{\fill}}lccccc}
                \toprule
                \multicolumn{1}{c}{\ } & \multicolumn{2}{c}{\ } & \multicolumn{1}{c}{PASTIS} & \multicolumn{1}{c}{MTLCC (2016)} \\
                \cmidrule(r){4-5}
                Ablation & Description & Params (millions) & oA / mIoU & oA / mIoU \\
                \midrule
                \multirow{2}{*}{Encoder Downsampling} & $T_1 = \frac{T}{2}$ & 1.1 & 83.2 $\pm$ 0.01 / 63.4 $\pm$ 0.1 & 90.15 $\pm$ 0.01 / 77.4 $\pm$ 0.2 \\
                & $T_1, T_2, T_3 = T$ & 1.67 & 83.5 $\pm$ 0.1 / 64.4 $\pm$ 0.1 & 90.4 $\pm$ 0.1 / 78.6 $\pm$ 0.2 \\
                \multirow{2}{*}{Encoder Layers} & w/out Gated Conv & 1.16 & 83.4 $\pm$ 0.1 / 64.1 $\pm$ 0.2 & 90.2 $\pm$ 0.1 / 78 $\pm$ 0.1 \\
                & Squeeze \& Excitation \cite{hu_squeeze-and-excitation_2018} & 1.26 & 83.5 $\pm$ 0.1 / 64.7 $\pm$ 0.2 & 90.3 $\pm$ 0.1 / 78.6 $\pm$ 0.1 \\
                \multirow{2}{*}{Decoder Layers} & Max Pool & 0.9 & 83.3 $\pm$ 0.04 / 64.2 $\pm$ 0.2 & 90.3 $\pm$ 0.1 / 78.2 $\pm$ 0.1 \\
                & Conv Transpose & 2.37 & 83.7 $\pm$ 0.1 / 64.7 $\pm$ 0.1 & 90.5 $\pm$ 0.1 / 78.7 $\pm$ 0.2 \\
                \bottomrule \addlinespace[0.3em]
                VistaFormer Performance & & 1.2 & 83.6 $\pm$ 0.1 / 64.8 $\pm$ 0.2 & 90.4 $\pm$ 0.1 / 78.7 $\pm$ 0.3 \\
                \bottomrule
        \end{tabular}
    }
    \vspace{1mm}
    \caption[VistaFormer Ablation Analysis]{We present the ablation analysis results for both the PASTIS and MTLCC benchmarks for semantic segmentation. For the MTLCC benchmark, we include the unknown class and for PASTIS we use fold-1 from the PASTIS benchmark which uses folds 1, 2, and 3 for training, fold 4 for validation; and fold 5 for testing. In keeping with Results in Section \ref{sec:results}, we report the mean and standard deviation for the mIoU and oA scores over three trials for the chosen PASTIS split and the MTLCC dataset.}
    \label{tab:ablation}
\end{table*}

\begin{figure*}[b]
    \centering
    \begin{subfigure}[b]{.48\textwidth}
        \centering
        \includegraphics[width=\linewidth]{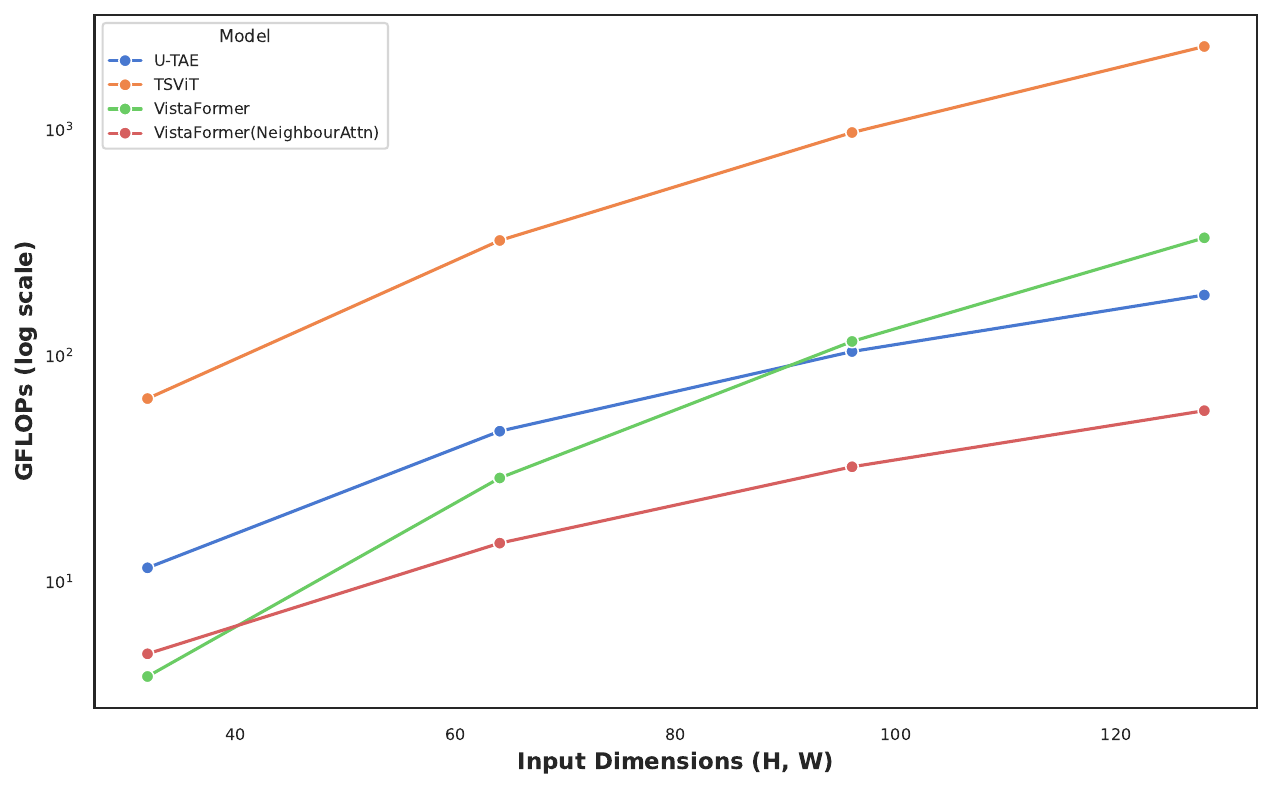}
        \caption{}
    \end{subfigure}
    \begin{subfigure}[b]{.48\textwidth}
        \centering
        \includegraphics[width=\linewidth]{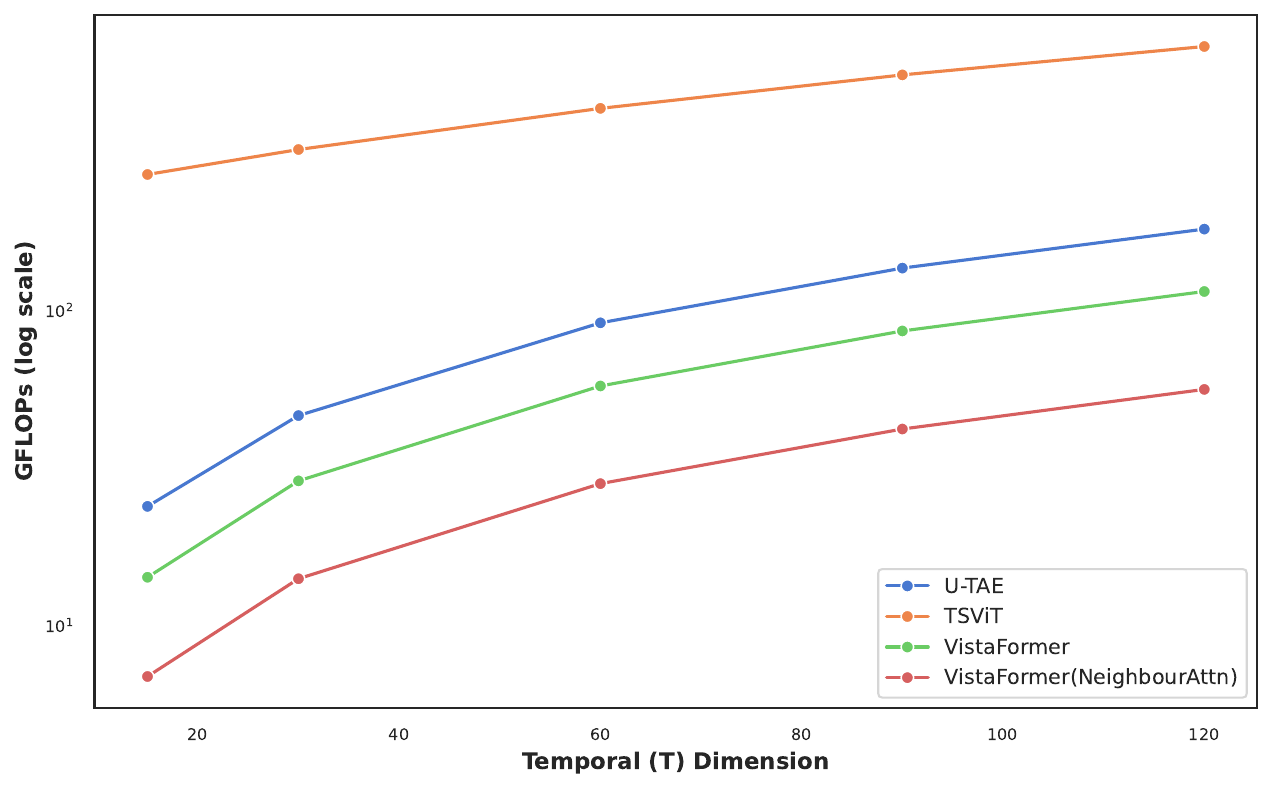}
        \caption{}
    \end{subfigure}
    \caption[SITS Model GFLOPs Comparison]{\textbf{(a)} Shows the scaling of floating point operations in GFLOPs of VistaFormer with MHSA and NA respectively compared with TSViT and U-TAE models using an input dimension of $(B, C, T, H, W) = (4, 10, 30, x_i, x_i)$ where we scale height and width dimensions using $x_i$. \textbf{(b)} Reflects the scaling of VistaFormer in terms of GFLOPs using input dimensions $(B, C, T, H, W) = (4, 10, t_i, 64, 64)$ where we scale $x_i$.}
    \label{fig:model_scalability}
\end{figure*}

\subsection{Ablations \label{sec:ablations}}

We report ablations concerning (a) decoder layers, (b) encoder layers, and (c) encoder downsampling. We find that using a max pooling layer instead of a 1D convolution in the decoder decreases model performance only slightly across all results excluding overall accuracy for the MTLCC dataset. We find that replacing trilinear interpolation with a transposed 3D convolution improved results inconsistently across datasets while requiring a considerable number of trainable parameters relative to the model size.

For encoder layers, we use gated convolutions by default to reduce input noise and find that not including this layer consistently results in a consistent decrease in both oA and mIoU scores. We introduced a Squeeze and Excitation (SE) \cite{hu_squeeze-and-excitation_2018} layer as an additional convolution filtering mechanism in each of the downsampling encoder layers, though we found inconsistent results in our tests and omitted the layer from the chosen architecture to preserve simplicity. This layer was used as a potential noise reduction technique since it adaptively recalibrates channel-wise feature responses which can reduce noise by emphasizing important input features and suppressing irrelevant ones.

Concerning encoder downsampling, the base model in Figure \ref{fig:vistaformer} in effect uses a 2D convolution in the first layer, only downsampling the height and width in keeping with \cite{garnot_panoptic_2021}. We find that both downsampling $T$ in the first encoder, giving us $T_1 = \frac{T}{2}$, and not downsampling $T$ in any encoder layer, resulted in decreased model performance relative to our proposed model. These results indicate that downsampling the temporal dimension can be performed effectively for SITS data, contrary to the preservation of the temporal dimension used in \cite{garnot_panoptic_2021, tarasiou_vits_2023}, allowing our model to reduce the sequence length used in self-attention layers, and subsequently the complexity of the model.

\subsection{Model Scalability \label{sec:scalability}}

VistaFormer treats each sequence entry in the temporal dimension $T$ as a unique sample which increases the floating point operations performed, since we compute the attention for each sequence of length $H \times W$, for each of the $T$ samples, and we are careful to limit the downsampling used at each layer of the encoder block. To address this, we demonstrate that substituting MHSA with NA dramatically improves the model's scalability relative to TSViT by reducing the spatial dimension used for computing self-attention. More specifically, we find that 2D Neighbourhood attention dramatically reduces the number of floating point operations required both for small input dimensions and as the spatial dimensions increase as seen in Figure \ref{fig:model_scalability}.

\section{Conclusion}

We have demonstrated a lightweight SITS semantic segmentation model that achieves efficiency by (a) downsampling both spatial and temporal dimensions (b) employing gated convolutions to boost model performance without pre-training for masking out clouds or atmospheric distortions, (c) using position-free self-attention layers to simplify the architecture, and (d) proposing a lightweight decoder to reduce computational complexity. Further, the position-free self-attention layers make this model extensible and simpler to use than existing models. We find that VistaFormer outperforms the current state-of-the-art model, TSViT, in terms of oA and mIoU performance while using a small fraction of the number of floating point operations and fewer trainable parameters. The ablation analysis highlights the importance of carefully selecting downsampling strategies and maintaining simplicity in the model architecture to achieve optimal performance.

We anticipate the model's efficiency and straightforward design will provide immediate benefits for remote sensing researchers and practitioners, particularly those with limited computational resources looking to train deep learning models. Our model is designed to be especially simple to train and deploy while offering considerable improvements over compared models. While there is a risk that semantic segmentation models could be used improperly to identify objects with the intent to cause harm, we are confident that the positive outcomes of making this model available will be significantly greater, since identifying crop types plays an important role in ensuring food security and creating climate adaptation strategies.

Some of the approaches in this model are generalizable. Given the effectiveness of gated convolutions for improving our model results across both benchmarks, we believe that mechanisms like partial convolutions \cite{liu_image_2018} or pre-training gated convolutions using cloud masks may be of benefit for filtering noise found in remote sensing inputs. Similarly, our proposed architecture introduces lightweight design patterns that can be adapted for different image time series segmentation tasks. Building on the strengths of our current SITS semantic segmentation model, several follow-up research tasks are proposed to expand its capabilities and refine its performance. The model could be extended to be applied to panoptic segmentation to further investigate the model's versatility; and experimentation with additional attention mechanisms like deformable attention, window attention, and additional configurations for neighbourhood attention. 

These proposed tasks could enhance the model's current capabilities and extend the applicability of the proposed architecture to more complex and demanding real-world applications. By continuing to explore these avenues, we hope to push the boundaries of SITS semantic segmentation, ultimately paving the way for more adaptable, efficient, and robust models that address increasingly complex real-world challenges.

 {\small
\bibliographystyle{ieee_fullname}
\bibliography{references}
}
% \printbibliography % Prints bibliography

\end{document}